\documentclass[10pt,twocolumn,letterpaper]{article}

\usepackage{wacv}              % WACV CAMERA-READY version; preserves visible authors from the source
\usepackage{amssymb}
\usepackage{algorithm}
\usepackage{algpseudocode}
\usepackage{multirow}
\usepackage{graphicx}
\usepackage{placeins}
\usepackage{dblfloatfix}
\newcommand{\xmark}{$\times$}

\definecolor{wacvblue}{rgb}{0.21,0.49,0.74}
\usepackage[pagebackref,breaklinks,colorlinks,allcolors=wacvblue]{hyperref}
\usepackage{xcolor}
\definecolor{languagegray}{RGB}{153,76,0}
\def\wacvPaperID{2968} % *** Enter the WACV Paper ID here
\def\confName{WACV}
\def\confYear{2027}
\hypersetup{hypertexnames=false}
\title{Temporally Ordered Region-Token Mamba with Logit-Space Diffusion for Remote Sensing Change Detection}

\author{
Maneet Chatterjee\textsuperscript{1}\thanks{Equal contribution.}\quad
Aparup Ghosh\textsuperscript{1}\footnotemark[1]\quad
Anuvab Sen\textsuperscript{2}$^\dagger$\quad
Udayon Sen\textsuperscript{3}\quad
Arnav Aditya\textsuperscript{4}\quad
Yixin Zhang\textsuperscript{5}$^\dagger$\\[0.5em]
\textsuperscript{1}Indian Institute of Engineering Science and Technology (IIEST), Shibpur\\
\textsuperscript{2}Georgia Institute of Technology\\
\textsuperscript{3}Nanyang Technological University\\
\textsuperscript{4}Indian Institute of Technology, Delhi,\quad
\textsuperscript{5}Duke University
}

\begin{document}
\maketitle

\newcommand{\blfootnote}[1]{%
  \begingroup
  \renewcommand\thefootnote{}\footnote{#1}%
  \addtocounter{footnote}{-1}%
  \endgroup
}
\blfootnote{$^\dagger$Corresponding authors: \href{mailto:asen74@gatech.edu}{asen74@gatech.edu} and \href{mailto:yz696@duke.edu}{yz696@duke.edu}}
\blfootnote{Code: \url{https://github.com/Aparup2139/BMD-CD_Remote_Sensing_Change_Detection}}

\begin{abstract}
Change Detection (CD) in remote sensing requires jointly modeling bitemporal context and fine-grained spatial details to accurately localize changed regions. While dense attention effectively captures bitemporal relationships, its computational cost limits scalability to high-resolution imagery. In contrast, conventional feature fusion and coarse decoding often struggle to distinguish true changes from appearance variations and preserve object boundaries. To overcome these limitations, we propose Bitemporal Mamba-Diffusion for Change Detection (BMD-CD), a framework that integrates temporally structured state-space modeling with logit-space diffusion refinement. BMD-CD first converts deep bitemporal features into region tokens and organizes them into explicit temporal partitions for bidirectional state-space propagation. It then employs a Learnable Trigonometric Fusion (LTF) module to integrate temporal information and a Conditional Diffusion Decoder that refines coarse predictions in logit space with a short DDIM schedule to generate fine-grained, high-resolution change maps. Extensive experiments on LEVIR-CD, WHU-CD, DSIFN-CD, CDD, and S2Looking demonstrate the effectiveness of the proposed approach. BMD-CD achieves IoU scores of 0.923, 0.957, 0.882, and 0.979 on WHU-CD, DSIFN-CD, LEVIR-CD, and CDD, respectively, and further exhibits strong zero-shot robustness on ValaisCD and B-FLAIR-test.
\end{abstract}

\section{Introduction}
Change detection (CD) aims to identify meaningful changes between remote-sensing images of the same geographical area acquired at different times. It supports a wide range of applications, including tracking urban expansion~\cite{levircd}, assessing disaster damage~\cite{gupta2019creating}, and monitoring changes in natural ecosystems.
Despite significant advances, reliable change detection remains challenging for three reasons. First, variations in illumination, seasonal conditions, and sensor characteristics can introduce pseudo-changes, producing substantial appearance differences between bitemporal images that do not correspond to meaningful changes on the ground. Second, repeated downsampling during feature extraction can erode fine spatial details, leading to boundary degradation, particularly for thin or irregularly shaped regions. Third, high-resolution image pairs require the simultaneous modeling of local details and long-range cross-temporal dependencies, making efficient feature interaction difficult. An effective change detection model must therefore suppress non-target variations, preserve fine object boundaries, and capture global cross-temporal context without incurring excessive computational costs.
Existing methods have individually investigated pseudo-change suppression, boundary preservation, and global cross-temporal modeling through advances in feature comparison, multiscale decoding, and context modeling. While these approaches have achieved promising results, they are often designed independently, resulting in architectures that excel at one aspect of the problem while providing limited support for others. Consequently, a unified solution that simultaneously achieves robustness to pseudo-changes, precise boundary localization, and efficient structured cross-temporal reasoning remains needed. To address this need, we propose Bitemporal Mamba-Diffusion for Change Detection (BMD-CD), a unified framework that integrates structured temporal interaction, learnable feature fusion, and targeted prediction refinement through three complementary components:
\begin{enumerate}
\item The \textbf{Bitemporal Ordered Mamba Operator (BOMO)} organizes region tokens into explicit temporal partitions, with all $t_1$ tokens preceding all $t_2$ tokens, and performs bidirectional state-space propagation for global cross-temporal interaction with linear complexity.

\item The \textbf{Learnable Trigonometric Fusion (LTF)} module combines two feature representations using sine and cosine weights parameterized by learned angles, with each angle shared by two consecutive channel positions.

\item The \textbf{Conditional Diffusion Decoder (CDD)} refines coarse logits directly in logit space with a five-index DDIM schedule (four reverse transitions), improving the delineation of thin and irregularly shaped regions around ambiguous boundaries.

\end{enumerate}
We evaluate BMD-CD on five benchmark change-detection datasets and assess its zero-shot transferability to ValaisCD \cite{twoplayer} and B-FLAIR-test \cite{bflair}. We further compare its detection accuracy and inference efficiency against existing state-of-the-art methods. The results demonstrate that BMD-CD achieves competitive performance while maintaining a favorable accuracy-efficiency trade-off, as illustrated in Tab.~\ref{tab:efficiency}.

%----------------------------------------------------------------------
\section{Related Work}

\subsection{Robust Bitemporal Representation}
Early Siamese CNN-based change detection networks commonly extract features from the two temporal images using shared encoders; change representations are then constructed through feature differencing, concatenation, or fusion~\cite{fcsiam2018}. DDPM-CD~\cite{bandara2022ddpm} explores diffusion-based representation learning to improve robustness. It processes each temporal image at multiple selected diffusion timesteps and extracts intermediate representations from several post-bottleneck decoder levels at each timestep. These multilevel, multi-timestep features are then aggregated and supplied to a lightweight change classifier. While the resulting representations exhibit robustness to environmental perturbations, change-sensitive and change-invariant information are not explicitly organized into separate branches. Our LTF module instead combines two learned feature representations using sine and cosine weights, with each trainable mixing angle shared by two consecutive channel positions.

\subsection{Boundary Prediction Refinement}
Accurate boundary localization requires awareness of high-level semantic information while retaining fine spatial details. CNN-based encoder-decoder methods use skip connections, dense connections, and multiscale feature fusion to help recover information lost during downsampling and improve boundary reconstruction~\cite{fang2021snunet}. More recent approaches strengthen dense representations using larger pretrained transformer backbones~\cite{next2former2024} or foundation segmentation models such as SAM2-CD~\cite{qin2025sam2cd}.
Generative models provide an alternative approach: RemoteVAR~\cite{remotevar2024} formulates change-map prediction as coarse-to-fine autoregressive generation, whereas DDPM-CD~\cite{bandara2022ddpm} uses its diffusion model mainly as a feature extractor, with final change predictions produced by a lightweight classification head on concatenated features. In contrast, our Conditional Diffusion Decoder (CDD) performs diffusion directly in the logit space, employing a small number of DDIM sampling steps~\cite{ddim2021} to refine coarse change maps, resolve boundary ambiguities, and improve localization accuracy with minimal computational overhead.

\subsection{Efficient Global and Cross-Temporal Modeling}
While CNN-based methods efficiently extract local and multi-scale features, they often struggle to capture long-range dependencies~\cite{fcsiam2018,fang2021snunet,zhang2024convolutional}. Transformer-based methods, such as BIT~\cite{bit2022} and ChangeFormer~\cite{changeformer2022}, improve global and cross-temporal interaction through attention mechanisms. However, the computational and memory costs of global self-attention scale quadratically with the number of tokens, limiting its efficiency on high-resolution imagery.

Selective state-space models, particularly Mamba~\cite{gu2023mamba}, offer linear sequence complexity~\cite{gu2022s4} while supporting long-range information propagation. Vision adaptations serialize two-dimensional feature maps into one-dimensional scan sequences~\cite{zhu2024vision,liu2024vmamba}. In remote sensing, RSMamba demonstrates the potential of state-space modeling for scene classification~\cite{rsmamba2024}, while EEMamba investigates hardware-aware, energy-efficient state-space modeling for satellite image classification~\cite{chatterjee2025eemamba}.
For bitemporal change detection, ChangeMamba, MF-VMamba, CDMamba, and M-CD adapt state-space models through dedicated spatial, multiscale, and temporal interaction mechanisms~\cite{chen2024changemamba,
zhang2024mfvmamba,zhang2024cdmamba,mcd2026}. Related linear-time recurrent alternatives have also been explored through RWKV-based change detection~\cite{yang2026changerwkv}. Our approach differs from these methods by organizing region tokens into deterministic temporal partitions before bidirectional state-space propagation.

\section{Method}
\begin{figure*}
    \centering
    \includegraphics[width=\linewidth]{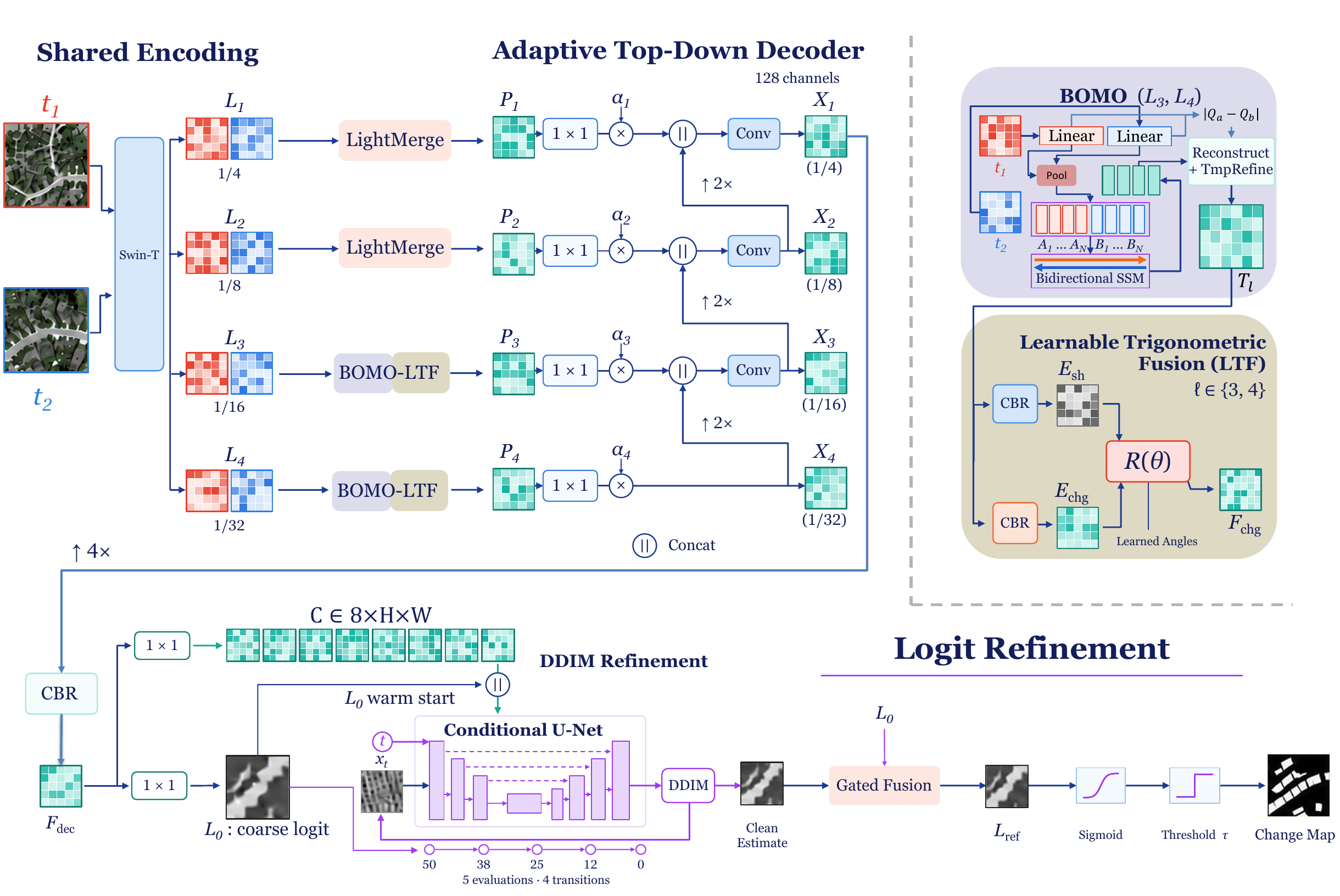}
    \caption{Overview of BMD-CD. A shared Swin-T backbone extracts four levels of bitemporal features. LightMerge fuses shallow features, while BOMO processes deep features through temporally ordered region tokens, bidirectional state-space propagation, and difference-based refinement. LTF combines two learned representations using sine and cosine weights. An adaptive top-down decoder integrates the resulting features to produce coarse logits $L_0$ and an eight-channel condition map $C$. Conditioned on both $L_0$ and $C$, a diffusion decoder performs DDIM refinement initialized from $L_0$. The refined estimate is gated with $L_0$, followed by sigmoid activation and thresholding to produce the final change map.}
    \label{fig:model}
\end{figure*}

As shown in Fig.~\ref{fig:model}, BMD-CD contains four stages:

\begin{enumerate}
\item A shared Swin-T~\cite{liu2021swin} backbone extracts four-level features from both images.
\item LightMerge processes the two shallow levels, while the BOMO--LTF branch models the two deep levels.
\item A top-down decoder fuses the hierarchy and produces coarse logits $\mathbf{h}$ and diffusion condition features $\mathbf{C}$.
\item A conditional decoder applies a five-index DDIM schedule to refine $\mathbf{h}$ in logit space.
\end{enumerate}

\subsection{Multi-hierarchy Patch Encoding}
Given a bitemporal pair $(I_a,I_b)$, the shared Swin-T backbone produces $\{f_l^a,f_l^b\}_{l=1}^{4}$ at progressively larger receptive fields but coarser resolutions. Shallow features retain local detail and are processed by LightMerge at $l\in\{1,2\}$; the deeper levels $l\in\{3,4\}$ are passed to the BOMO--LTF branch described next.

\subsection{LightMerge for Shallow Features}

At levels $l\in\{1,2\}$, LightMerge combines absolute-difference and element-wise co-activation cues~\cite{fcsiam2018,fang2021snunet}. Consecutive $1\times1$ and $3\times3$ Conv--BN--ReLU blocks fuse these cues, after which a Squeeze-and-Excitation (SE)~\cite{hu2018senet} block reweights the channels:
\begin{equation}
\mathbf{x}_l
=
\operatorname{SE}\!\left(
\operatorname{CBR}_{3\times3}\!\left(
\operatorname{CBR}_{1\times1}\!\left(
\left[|f_l^a-f_l^b|;f_l^a\odot f_l^b\right]
\right)\right)\right).
\end{equation}
A mean--max spatial attention map~\cite{woo2018cbam} then suppresses background responses and produces the shallow output
\begin{equation}
\mathbf{P}_l
=
\mathbf{x}_l\odot
\sigma\!\left(
\operatorname{BN}\!\left(
\operatorname{Conv}_{7\times7}\!\left(
[\operatorname{Mean}_c(\mathbf{x}_l);\operatorname{Max}_c(\mathbf{x}_l)]
\right)\right)\right).
\end{equation}
Here, $\operatorname{CBR}$ denotes convolution, batch normalization, and ReLU. The resulting features retain the local and boundary information used by the top-down decoder.

\subsection{BOMO--LTF Branch for Deep Features }
\label{sec:bomo}

Deep features encode larger structures and therefore benefit from broader cross-temporal context than the pointwise operations used by LightMerge. Although Mamba has linear sequence complexity~\cite{gu2023mamba}, scanning both feature maps directly would still require $2H_lW_l$ state updates. BOMO instead pools each map into $K=64$ region tokens, performs bidirectional propagation over the resulting $2K$-token sequence, and reconstructs a spatial feature. Temporal refinement and LTF then form the change-oriented representation supplied to the decoder.
\begin{algorithm}[t]

\caption{BOMO and Temporal Refinement at Deep Hierarchy Level \(l\)}
\label{alg:bomo}
\small
\textbf{Input:} $f_l^a,f_l^b\in\mathbb{R}^{B\times C_l\times H_l\times W_l}$, $l\in\{3,4\}$\\
\textbf{Output:} $\mathbf{T}_l\in\mathbb{R}^{B\times C_l'\times H_l\times W_l}$
\begin{algorithmic}[1]
\Statex \Comment{\textbf{Region-token construction}}
\State $q_l^a\leftarrow W_l^a f_l^a,\quad q_l^b\leftarrow W_l^b f_l^b$
\label{line:bomo-projection}
\State $\mathbf{N}_l^a\leftarrow\mathcal{P}_K(q_l^a),\quad
       \mathbf{N}_l^b\leftarrow\mathcal{P}_K(q_l^b)$
\State $\mathbf{S}_l\leftarrow\mathrm{Cat}_{\mathrm{tok}}(\mathbf{N}_l^a,\mathbf{N}_l^b)$
\label{line:bomo-sequence}
\Statex \Comment{\textbf{Bidirectional state-space propagation}}
\State $(\mathbf{X}_l,\mathbf{G}_l)\leftarrow
       \mathrm{Split}_{C}(W_{\mathrm{in}}\mathbf{S}_l)$
\State $\widehat{\mathbf{X}}_l\leftarrow\mathrm{SiLU}(\mathbf{X}_l)$
\State $\Delta_l\leftarrow
       \mathrm{Clamp}(\mathrm{Softplus}(W_{\delta}\widehat{\mathbf{X}}_l),10^{-4},1)$
\State $\mathbf{H}_l^{\rightarrow}\leftarrow
       \mathrm{SSM}(\widehat{\mathbf{X}}_l,\Delta_l)$
\State $\mathbf{H}_l^{\leftarrow}\leftarrow\mathrm{Rev}\left(
       \mathrm{SSM}(\mathrm{Rev}(\widehat{\mathbf{X}}_l),
                    \mathrm{Rev}(\Delta_l))\right)$
\State $\mathbf{Z}_l\leftarrow
       (\mathbf{H}_l^{\rightarrow}+\mathbf{H}_l^{\leftarrow})
       \odot\mathrm{SiLU}(\mathbf{G}_l)$
\State $\mathbf{Y}_l\leftarrow
       \mathrm{LN}(\mathbf{S}_l+W_{\mathrm{out}}\mathbf{Z}_l)$
\label{line:bomo-state}
\Statex \Comment{\textbf{Spatial and temporal reconstruction}}
\State $\mathbf{Y}_l^b\leftarrow\mathbf{Y}_l[:,K:2K,:]$
\State $\mathbf{U}_l\leftarrow
       \mathrm{SE}\!\left(\mathrm{CBR}_{3\times3}\!\left(
       \mathrm{Up}(\mathrm{Reshape}(\mathbf{Y}_l^b))\right)\right)$
\State $\mathbf{D}_l\leftarrow|q_l^a-q_l^b|,\quad
       \mathbf{M}_l\leftarrow\mathbf{D}_l+\gamma_l(\mathbf{U}_l-\mathbf{D}_l)$
\label{line:bomo-reconstruction}
\State $\widetilde{\mathbf{D}}_l\leftarrow|V_l f_l^a-V_l f_l^b|$
\State $\mathbf{A}_l\leftarrow\sigma\!\left(
       \mathrm{BN}\!\left(\mathrm{Conv}_{1\times1}
       (\mathrm{Cat}_{C}(\mathbf{M}_l,\widetilde{\mathbf{D}}_l))\right)\right)$
\State $\mathbf{T}_l\leftarrow
       \mathbf{A}_l\odot\mathbf{M}_l+
       (1-\mathbf{A}_l)\odot\widetilde{\mathbf{D}}_l$
\label{line:bomo-refinement}
\end{algorithmic}
\end{algorithm}
\subsubsection{Ordered Region Tokens and State Propagation}

As summarized in Alg.~\ref{alg:bomo}, the independent projections in line~\ref{line:bomo-projection} map both observations to $C_l'$ channels. The operator $\mathcal{P}_K$ then adaptively averages each map to an $8\times8$ grid and flattens it in raster order, yielding $\mathbf{N}_l^a,\mathbf{N}_l^b\in\mathbb{R}^{B\times K\times C_l'}$ with $K=64$. Corresponding token indices summarize the same spatial regions. Line~\ref{line:bomo-sequence} concatenates all $I_a$ tokens before all $I_b$ tokens to form $\mathbf{S}_l\in\mathbb{R}^{B\times2K\times C_l'}$; this fixed temporal arrangement gives BOMO its \textit{ordered} designation and replaces $2H_lW_l$ spatial states with 128 region states.

The state-space block splits $\mathbf{S}_l$ into content and gating branches and predicts bounded, input-dependent transition steps. The same selective SSM scans the fixed sequence forward and backward, after which its gated outputs are combined with the input through a normalized residual connection (line~\ref{line:bomo-state}). Because the $I_b$ partition follows all $I_a$ tokens, its forward states contain context propagated from the first observation; the reverse scan supplies complementary context from the opposite direction. Both scans remain linear in the compact sequence length.

\subsubsection{Spatial Reconstruction and Temporal Refinement}

The final $K$ positions of $\mathbf{Y}_l$ correspond to $I_b$ and contain context propagated over the ordered sequence. BOMO reshapes this partition to an $8\times8$ grid, upsamples it to $H_l\times W_l$, and applies local convolution and SE calibration to obtain $\mathbf{U}_l$. Because region pooling can weaken precise spatial cues, line~\ref{line:bomo-reconstruction} combines this context with the projected difference $\mathbf{D}_l=|q_l^a-q_l^b|$. The scalar $\gamma_l$, initialized to zero, makes BOMO begin from the direct difference and learn a contextual correction $\mathbf{M}_l$.

A spatially and channel-wise adaptive gate then balances $\mathbf{M}_l$ against a second localized difference $\widetilde{\mathbf{D}}_l=|V_lf_l^a-V_lf_l^b|$. As shown in line~\ref{line:bomo-refinement}, larger values of $\mathbf{A}_l$ favor the reconstructed BOMO context, whereas smaller values favor the direct difference. The resulting $\mathbf{T}_l$ is passed to LTF.
\subsubsection{Learnable Trigonometric Fusion (LTF)}
LTF combines two candidate representations of the refined feature
$\mathbf{T}_l$. Two independent Conv--BN--ReLU blocks produce
$\mathbf{E}_l^{(1)}=\phi_l^{(1)}(\mathbf{T}_l)$ and
$\mathbf{E}_l^{(2)}=\phi_l^{(2)}(\mathbf{T}_l)$, each containing
$C_l'$ channels. Corresponding channels across the two branches
are fused using sine and cosine weights. Every two consecutive
channel positions share one learned angle, giving
$J_l=C_l'/2$ angles in total. Specifically, angle $\theta_{l,j}$
controls fusion at channel positions $2j-1$ and $2j$:
\begin{equation}
\begin{aligned}
\mathbf{F}_{l,2j-1}^{\mathrm{chg}}
&=
\sin\theta_{l,j}\,\mathbf{E}_{l,2j-1}^{(1)}
+
\cos\theta_{l,j}\,\mathbf{E}_{l,2j-1}^{(2)},\\
\mathbf{F}_{l,2j}^{\mathrm{chg}}
&=
\sin\theta_{l,j}\,\mathbf{E}_{l,2j}^{(1)}
+
\cos\theta_{l,j}\,\mathbf{E}_{l,2j}^{(2)}.
\end{aligned}
\end{equation}
During training, an auxiliary loss encourages variation among
the mixing angles at levels $3$ and $4$ and consistency between
shallow bitemporal features over unchanged pixels:
\begin{equation}
\begin{aligned}
\mathcal{L}_{\mathrm{aux}}
&=
\sum_{l\in\{3,4\}}
\frac{1}{\operatorname{Var}(\boldsymbol{\theta}_l)+0.1}
\\
&\quad+
0.1\,\mathcal{L}_1\!\left(
f_1^a\odot\mathbf{U}_{\mathrm{unch}},
f_1^b\odot\mathbf{U}_{\mathrm{unch}}
\right),
\end{aligned}
\end{equation}
where $\boldsymbol{\theta}_l$ collects the learned angles at
level $l$, $f_1^a$ and $f_1^b$ are the shallow features from the
two temporal inputs, $\mathbf{U}_{\mathrm{unch}}$ is the resized
complement of the ground-truth change mask, $\odot$ denotes
elementwise multiplication, and $\mathcal{L}_1$ denotes mean
absolute error. The auxiliary loss enters the overall training
objective with weight $0.1$.
\subsection{Hierarchical Fusion and Coarse Logit Projection}
\label{sec:hierarchical_fusion}

The decoder receives the LightMerge outputs $\mathbf{P}_1,\mathbf{P}_2$ and the change-oriented LTF outputs $\mathbf{P}_3=\mathbf{F}_3^{\mathrm{chg}}$, $\mathbf{P}_4=\mathbf{F}_4^{\mathrm{chg}}$. Level-specific $1\times1$ projections map them to $D=128$ channels. Global descriptors of the resolution-aligned features determine sample-dependent scale weights:
\begin{equation}
\begin{aligned}
\widetilde{\mathbf{P}}_l&=\phi_l^{\mathrm{lat}}(\mathbf{P}_l),
&\mathbf{a}_l&=\operatorname{GAP}\!\left(\mathcal{R}(\widetilde{\mathbf{P}}_l)\right),\\
\boldsymbol{\alpha}&=\operatorname{Softmax}\!\left(\operatorname{MLP}([\mathbf{a}_1;\cdots;\mathbf{a}_4])\right),
&\overline{\mathbf{P}}_l&=\alpha_l\widetilde{\mathbf{P}}_l,
\end{aligned}
\end{equation}
where $\operatorname{GAP}$ denotes global average pooling. Inspired by feature pyramids~\cite{lin2017fpn}, fusion proceeds from the deepest level upward by concatenating each weighted lateral feature with the upsampled decoder state:
\begin{equation}
\begin{aligned}
\mathbf{X}_4&=\overline{\mathbf{P}}_4,\\
\mathbf{X}_l&=\Phi_l\!\left([\operatorname{Up}_l(\mathbf{X}_{l+1});\overline{\mathbf{P}}_l]\right),
\quad l\in\{3,2,1\},\\
\mathbf{F}_{\mathrm{dec}}&=\phi_{\mathrm{head}}(\operatorname{Up}(\mathbf{X}_1)),\\
\mathbf{h}&=W_{\mathrm{logit}}\mathbf{F}_{\mathrm{dec}},
\qquad \mathbf{C}=W_{\mathrm{cond}}\mathbf{F}_{\mathrm{dec}}.
\end{aligned}
\end{equation}
Each $\Phi_l$ contains two Conv--BN--ReLU blocks, and $\phi_{\mathrm{head}}$ produces a 64-channel full-resolution feature. The $1\times1$ output projections yield raw coarse logits $\mathbf{h}\in\mathbb{R}^{B\times1\times H\times W}$ and an eight-channel condition map $\mathbf{C}$ for diffusion.

\subsection{Conditional Diffusion Decoder in Logit Space}
\label{sec:diffusion}

\begin{figure}
    \centering
    \includegraphics[width=0.75\linewidth]{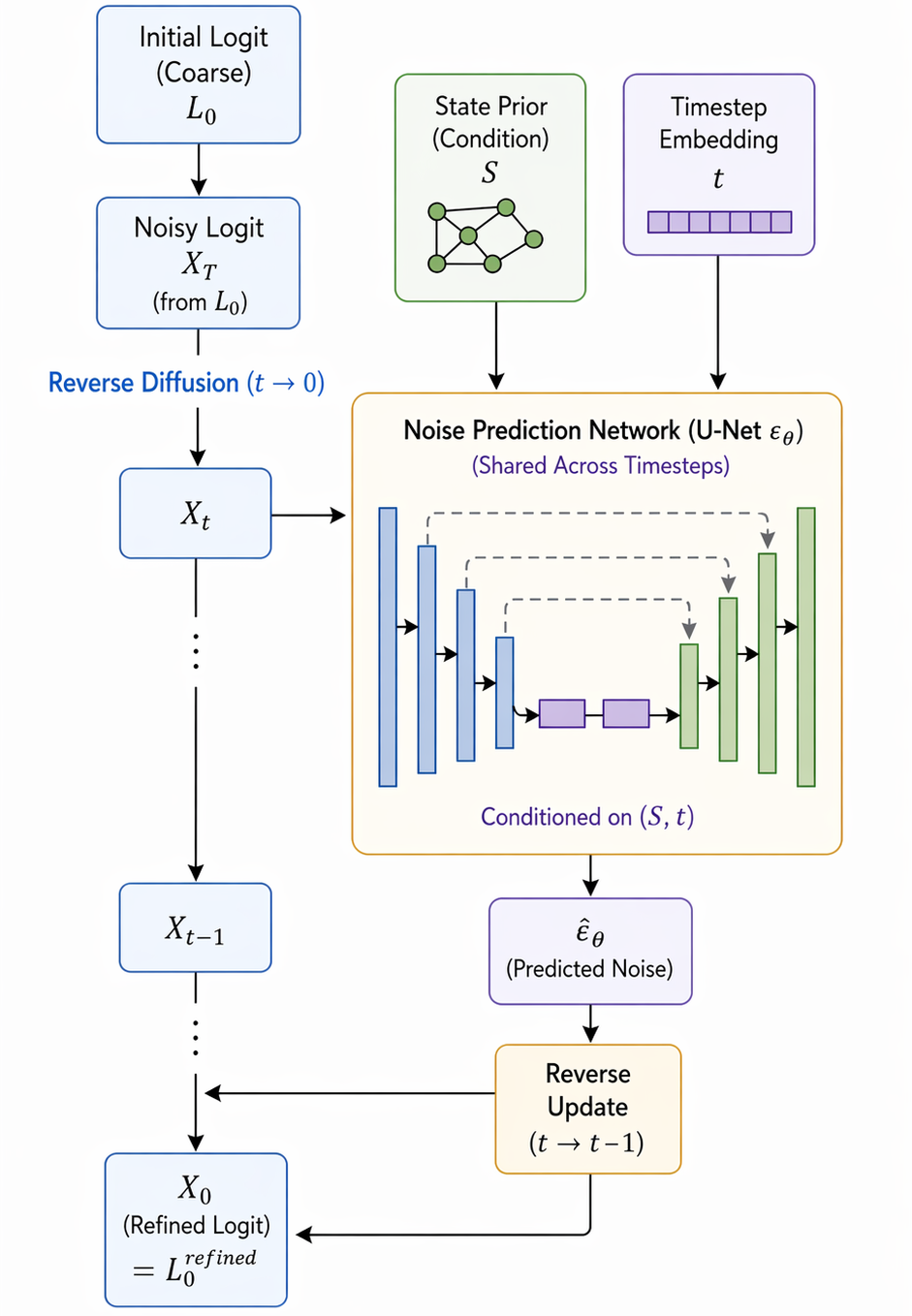}
    \caption{Architecture of Conditional Diffusion Decoding.}
    \label{fig:diffusion}
\end{figure}

As shown in Fig.~\ref{fig:diffusion}, the coarse logit map $L_0$ provides an initial change prediction, while the hierarchical context map $\mathbf{C}$ supplies multi-scale decoder information. We denote the coarse logit from Eq.~(6) by $L_0\equiv\mathbf{h}$ and the eight-channel diffusion condition by $S_{\mathrm{cond}}\equiv\mathbf{C} \in\mathbb{R}^{B\times8\times H\times W}$. The decoder is therefore a conditional logit-space refiner: it learns a residual correction around the coarse prediction rather than generating a change map from scratch. During training, the binary mask $Y$ is converted into a bounded clean logit target $Y^\ell$. The forward process corrupts $Y^\ell$, and the denoiser predicts the injected noise conditioned on $(x_t,t,L_0,S_{\mathrm{cond}})$. The recovered clean-logit estimate is fused with $L_0$ through a learned residual gate to produce the final refined logit $L_{\mathrm{ref}}$. BCE and Dice losses are applied to $L_{\mathrm{ref}}$, while the diffusion branch is additionally trained with a noise-regression loss. This ensures that segmentation gradients supervise the final refinement path rather than only the coarse prediction. Full equations and gradient-flow details are provided in Supp.~S3.2--S3.5.

At inference, $Y^\ell$ is unavailable, so the deterministic DDIM trajectory is
initialized from $L_0$. The five-index DDIM schedule uses indices $\{50,38,25,12,0\}$ and four reverse transitions. The reported implementation evaluates the denoiser at each scheduled index, including the terminal index $t=0$, and the reported FLOPs/latency count these five U-Net evaluations. Since $\bar\alpha_0=1$, the terminal evaluation is output-neutral under the clean-logit recovery equation; retaining it makes the reported cost conservative relative to an equivalent four-call implementation. The final estimate is fused with $L_0$ using the same learned residual gate, converted to probabilities, and thresholded using the validation-selected $\tau_{\mathrm{bin}}$. No morphological post-processing is used. Additional inference details are given in Supp.~S3.4--S3.5.
\section{Experiments}

\begin{table*}[t]
\caption{Quantitative comparison on WHU-CD, DSIFN-CD, LEVIR-CD and CDD datasets. F1 and IoU are reported as ratios, while OA is reported in percentage. Best results in \textbf{bold}, second-best underlined. “–” indicates unavailable comparable prior results or artifacts, with retraining being computationally infeasible. $^{\dagger}$Change-class F1 from the source; IoU derived as F1/(2$-$F1).
}
\label{tab:cd_results}
\centering

\scriptsize
\setlength{\tabcolsep}{3pt}
\renewcommand{\arraystretch}{1.08}

\begin{tabular}{c c ccc ccc ccc ccc}
\toprule

& & \multicolumn{3}{c}{WHU-CD} 
& \multicolumn{3}{c}{DSIFN-CD} 
& \multicolumn{3}{c}{LEVIR-CD} 
& \multicolumn{3}{c}{CDD} \\

\cmidrule(lr){3-5} \cmidrule(lr){6-8} \cmidrule(lr){9-11} \cmidrule(lr){12-14}

Method & Extra data 
& F1 & IoU & OA 
& F1 & IoU & OA 
& F1 & IoU & OA 
& F1 & IoU & OA \\

\midrule
\multicolumn{14}{c}{\textbf{CNN-based Methods}} \\

FC-Siam-conc \cite{fcsiam2018} & None 
& 0.798 & 0.665 & 98.5 
& 0.597 & 0.426 & 87.6 
& 0.837 & 0.720 & 98.5 
& 0.751 & 0.601 & 94.9 \\

SNUNet \cite{fang2021snunet} & None 
& 0.835 & 0.717 & 98.7 
& 0.662 & 0.495 & 87.3 
& 0.882 & 0.788 & 98.8 
& 0.839 & 0.721 & 96.2 \\

IFNet \cite{dsifncd} & IN1k 
& 0.834 & 0.715 & 98.8 
& 0.601 & 0.430 & 87.8 
& 0.881 & 0.788 & 98.9 
& 0.840 & 0.719 & 96.03 \\

\midrule
\multicolumn{14}{c}{\textbf{CNN + Attention Methods}} \\

DT-SCN \cite{dtscn2020} & IN1k 
& 0.914 & 0.842 & 99.3 
& 0.706 & 0.545 & 82.9 
& 0.877 & 0.781 & 98.8 
& 0.921 & 0.853 & 98.2 \\

STANet \cite{levircd} & IN1k 
& 0.823 & 0.700 & 98.5 
& 0.645 & 0.478 & 88.5 
& 0.873 & 0.774 & 98.7 
& 0.841 & 0.722 & 96.1 \\

\midrule
\multicolumn{14}{c}{\textbf{Transformer-based Methods}} \\

BIT \cite{bit2022} & IN1k 
& 0.905 & 0.834 & 99.3 
& 0.876 & 0.780 & 92.3 
& 0.893 & 0.807 & 98.92 
& 0.889 & 0.800 & 97.5 \\

ChangeFormer \cite{changeformer2022} & None 
& 0.886 & 0.795 & 99.12 
& 0.947 & 0.887 & 93.2 
& 0.904 & 0.825 & 99.0 
& 0.946 & 0.898 & 98.7 \\

SAM2-CD \cite{qin2025sam2cd} & SAM-pretrained
& 0.926 & 0.889 & \underline{99.77}
& -- & -- & -- 
& 0.919 & 0.855 & 99.19 
& -- & -- & -- \\

PeftCD \cite{dong2026peftcd} & IN1k 
& \underline{0.959} & \underline{0.921} & 99.6
& -- & -- & -- 
& \underline{0.923} & \underline{0.856} & \underline{99.2}
& \underline{0.985} & \underline{0.970} & \underline{99.63} \\

\midrule
\multicolumn{14}{c}{\textbf{Self-supervised Pretraining}} \\

SimSiam \cite{siamsiam2022} & IN1k, IBSD, GoogleEarth 
& 0.847 & 0.734 & -- 
& -- & -- & -- 
& 0.880 & 0.786 & -- 
& -- & -- & -- \\

MoCo-v2 \cite{mocov2} & IN1k, IBSD, GoogleEarth 
& 0.882 & 0.789 & -- 
& -- & -- & -- 
& 0.879 & 0.784 & -- 
& -- & -- & -- \\

SeCo \cite{seco2021} & IN1k, IBSD, GoogleEarth 
& 0.883 & 0.790 & -- 
& -- & -- & -- 
& 0.881 & 0.787 & -- 
& -- & -- & -- \\

SaDL-CD \cite{sadlcd2023} & IN1k, IBSD, GoogleEarth 
& 0.909 & 0.833 & -- 
& -- & -- & -- 
& 0.899 & 0.818 & -- 
& -- & -- & -- \\

\midrule
\multicolumn{14}{c}{\textbf{Diffusion-based Methods}} \\

DDPM-CD \cite{bandara2022ddpm} & GoogleEarth 
& 0.927 & 0.863 & 99.4 
& 0.967 & 0.913 & 97.1
& 0.909 & 0.833 & 99.1 
& 0.956 & 0.916 & 99.0 \\

DiffRegCD$^{\dagger}$ \cite{diffregcd2024} & GoogleEarth 
& 0.874 & 0.776 & 99.0 
& 0.900 & 0.818 & 96.7
& 0.872 & 0.773 & 98.7 
& -- & -- & -- \\

\midrule
\multicolumn{14}{c}{\textbf{Mamba-based Methods}} \\

RSMamba \cite{zhao2024rsmamba} & IN1k 
& 0.927 & 0.865 & 99.4 
& 0.965 & 0.913 & 97.0 
& 0.897 & 0.814 & 98.9 
& 0.943 & 0.902 & 98.8 \\

ChangeMamba \cite{chen2024changemamba} & IN1k 
& 0.925 & 0.861 & 99.4 
& 0.875 & 0.778 & 95.8 
& 0.902 & 0.821 & 99.0 
& 0.944 & 0.920 & 99.0 \\

CDMamba \cite{zhang2024cdmamba} & IN1k 
& 0.937 & 0.882 & 99.5 
& 0.966 & 0.914 & 97.0 
& 0.907 & 0.831 & 99.0 
& 0.960 & 0.919 & 99.1 \\

M-CD \cite{mcd2026} & IN1k 
& 0.953 & 0.911 & 99.6 
& \underline{0.970} & \underline{0.935} & \underline{98.9} 
& 0.921 & 0.850 & \underline{99.2} 
& 0.982 & 0.963 & 99.5 \\

\midrule
\multicolumn{14}{c}{\textbf{Boundary-focused Methods}} \\

LRNet \cite{lrnet2023} & IN1k 
& 0.925 & 0.861 & 99.47 
& -- & -- & -- 
& 0.911 & 0.836 & 99.10 
& -- & -- & -- \\

\midrule
\multicolumn{14}{c}{\textit{Ours}} \\

\textbf{BMD-CD} & IN1k
& \textbf{0.960} & \textbf{0.923} & \textbf{99.78}
& \textbf{0.978} & \textbf{0.957} & \textbf{99.12}
& \textbf{0.937} & \textbf{0.882} & \textbf{99.35}
& \textbf{0.990} & \textbf{0.979} & \textbf{99.64} \\

\bottomrule
\end{tabular}
\end{table*}

\begin{table}[t]
\caption{3-px Boundary-F1 on LEVIR-CD (L) and WHU-CD (W) is reported as a percentage for methods with publicly available baseline change maps or official checkpoints. Baseline F1 scores are taken from the corresponding papers, while BMD-CD results are averaged over three seeds.}
\label{tab:boundary_seed}
\centering
\scriptsize
\setlength{\tabcolsep}{4pt}
\renewcommand{\arraystretch}{1.05}

\begin{tabular}{l|cc|cc}
\toprule
Method & F1$_{\mathrm L}$ & B-F1$_{\mathrm L}$ & F1$_{\mathrm W}$ & B-F1$_{\mathrm W}$ \\
\midrule
DDPM-CD \cite{bandara2022ddpm} & 90.9 & 84.8 & 92.7 & 88.7 \\
M-CD \cite{mcd2026}   & 92.1 & 85.5 & 95.3 & 90.0 \\
LRNet \cite{lrnet2023}  & 91.1 & 86.8 & 92.5 & 90.5 \\
\midrule
\textbf{BMD-CD} & \textbf{93.7} & \textbf{87.7} & \textbf{96.0} & \textbf{91.4} \\
\bottomrule
\end{tabular}
\end{table}

\subsection{Datasets}

We evaluate BMD-CD on four standard change-detection benchmarks: \textit{LEVIR-CD}~\cite{levircd}, \textit{DSIFN-CD}~\cite{dsifncd}, \textit{WHU-CD}~\cite{whucd}, and \textit{CDD}~\cite{cdd}. We further evaluate on \textit{S2Looking}~\cite{S2Looking}, which contains pronounced viewpoint and appearance variations, and assess cross-dataset generalization by training only on S2Looking and evaluating zero-shot on \textit{ValaisCD}~\cite{twoplayer}, and \textit{B-FLAIR-test}~\cite{bflair}. Detailed dataset splits, preprocessing, and transfer protocols are provided in Supp.~Sec.~S7 and Supp.~Tab.~S12.

% 4.2 Implementation Details
\subsection{Implementation Details}

We use a \textbf{Swin Transformer-Tiny (Swin-T)} backbone~\cite{liu2021swin}, pretrained on ImageNet-1K, as the shared multi-scale encoder. Optimization is performed with \textbf{AdamW}~\cite{loshchilov2017decoupled} using $(\beta_1,\beta_2)=(0.9,0.999)$ and weight decay $10^{-4}$. A \textbf{OneCycleLR} schedule is used with maximum learning rate $3\times10^{-4}$, 10\% warmup, and cosine annealing. Models are trained for 125 epochs with batch size 8 using \textbf{automatic mixed precision (FP16)}. The overall objective is
\begin{equation}
\mathcal{L}=\mathcal{L}_{\mathrm{BCE}}+\mathcal{L}_{\mathrm{Dice}}+0.1\mathcal{L}_{\mathrm{aux}}+0.5\mathcal{L}_{\mathrm{diff}}    
\end{equation}

The number of bitemporal region tokens $K\!=\!64$ is empirically selected through the validation split and applied to the test set. For the diffusion decoder, we use 100 forward steps with a cosine noise schedule following DDPM~\cite{nichol2021improved}; inference follows the warm-started five-index DDIM schedule described in Sec.~\ref{sec:diffusion}. All BMD-CD experiments are conducted on a single \textbf{NVIDIA RTX 4090 GPU (24GB)}. We report F1-score, Intersection-over-Union (IoU), Overall Accuracy (OA), and Boundary F1 (B-F1) with a 3-pixel tolerance following~\cite{csurka2013good}.

% --------------------------------------------------------------------------
% 4.4 SOTA Comparison
% --------------------------------------------------------------------------
\subsection{Comparison with State-of-the-Art}

\begin{figure*}[!t]
    \centering
    \includegraphics[width=0.75\textwidth]{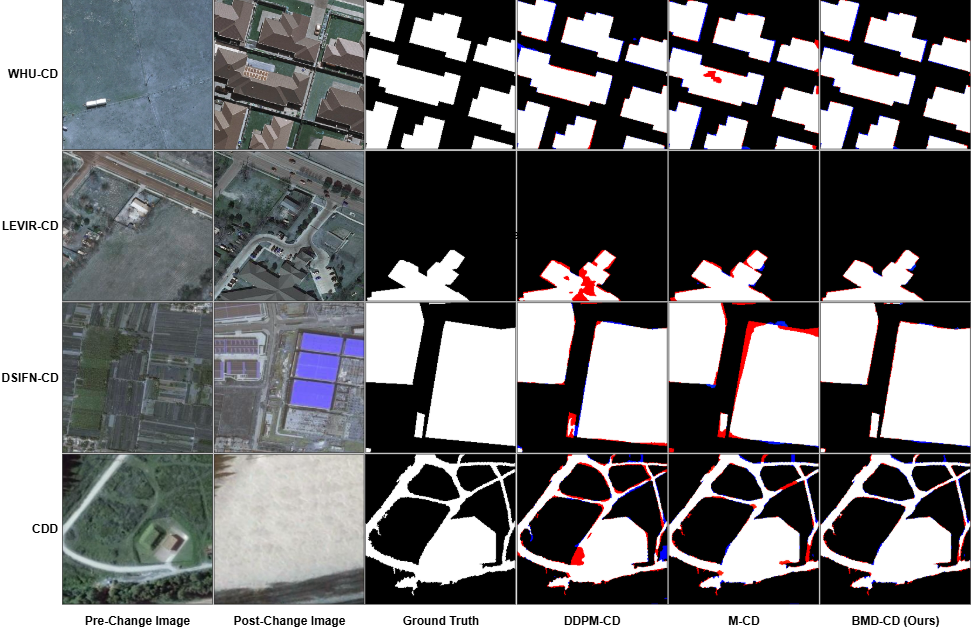}
    \caption{Change map comparison. The six columns show: Time 1 image, Time 2 image, Ground Truth, DDPM-CD, M-CD, and BMD-CD (Ours). In the error overlay, white denotes true positives (TP), black denotes true negatives (TN), red denotes false positives (FP), and blue denotes false negatives (FN).}
    \label{fig:qualitative}
\end{figure*}

\begin{table}[t]
\centering
\caption{Cross-dataset evaluation. BMD-CD is trained and tested on S2Looking~\cite{S2Looking}, also evaluated on ValaisCD and B-FLAIR-test datasets without target-domain training, fine-tuning, adaptation, validation, or threshold selection. ValaisCD baselines are included as protocol context and are not
directly comparable. Metrics are in \%; -- denotes not reported.}
\label{tab:zeroshot}

\scriptsize
\setlength{\tabcolsep}{2.2pt}
\renewcommand{\arraystretch}{1.05}

\resizebox{\columnwidth}{!}{%
\begin{tabular}{l|l|l|cccc}
\toprule
\textbf{Dataset} & \textbf{Method} & \textbf{Setting}
& \textbf{F1} & \textbf{Prec.} & \textbf{Rec.} & \textbf{IoU} \\
\midrule
\multirow{4}{*}{S2Looking}
& SChanger-base~\cite{schanger}
& Supervised
& 68.95 & 70.94 & 67.09 & --  \\

& ChangeStar2~\cite{changestar2}
& Supervised
& 67.80 & 69.2 & 66.50 & 51.3 \\

& LSKNet-S~\cite{lsknet}
& Supervised
& 67.52 & \textbf{71.90} & 63.64 & 50.96 \\

& \textbf{BMD-CD (ours)}
& \textbf{Supervised}
& \textbf{69.23} & 71.23 & \textbf{67.32} & \textbf{52.94} \\

\midrule
\multirow{3}{*}{ValaisCD}
& 2Player (BIT)~\cite{twoplayer}
& Unsupervised
& 19.00 & 11.00 & \textbf{73.00} & -- \\

& FC-Siam-Diff~\cite{fcsiam2018}
& Supervised, in-domain
& 46.00 & 41.00 & 52.00 & -- \\

& \textbf{BMD-CD (ours)}
& \textbf{Zero-shot}
& \textbf{49.40} & \textbf{50.20} & 48.63 & \textbf{32.80} \\

\midrule

\multirow{4}{*}{B-FLAIR-test \cite{bflair}}
& Dual UNet (STAR)~\cite{bflair}
& Zero-shot
& 75.90 & -- & -- & 61.10 \\

& Dual UNet (b-FLAIR)~\cite{bflair}
& Zero-shot
& 79.00 & -- & -- & 65.30 \\

& Dual UNet (FSC-180k)~\cite{bflair}
& Zero-shot
& \textbf{83.10} & -- & -- & \textbf{71.10} \\

& \textbf{BMD-CD (ours)}
& \textbf{Zero-shot}
& 76.30 & \textbf{75.10} & \textbf{77.54} & 61.68 \\

\bottomrule
\end{tabular}%
}

\vspace{1mm}
\end{table}
Tab.~\ref{tab:cd_results} reports quantitative comparisons across four benchmarks. Prior methods span CNNs~\cite{fcsiam2018,fang2021snunet}, attention models~\cite{levircd,bit2022,changeformer2022}, pretrained/foundation and adaptation methods~\cite{qin2025sam2cd,dong2026peftcd}, state-space models~\cite{zhao2024rsmamba,chen2024changemamba,zhang2024cdmamba,mcd2026}, and diffusion- or boundary-oriented designs~\cite{bandara2022ddpm,diffregcd2024,lrnet2023}. In contrast, BMD-CD jointly models cross-temporal interaction with diffusion-style logit-space residual refinement. The BOMO module performs temporally ordered cross-temporal propagation with linear complexity, while the diffusion decoder performs logit-space correction guided by global context. This combination improves region-level accuracy in the main comparison and boundary quality in the B-F1 evaluation. As shown in Tab.~\ref{tab:cd_results}, BMD-CD achieves competitive or best performance among entries with available comparable results, improving F1 and IoU over most CNN, transformer, state-space, and diffusion-based baselines under the reported protocols. We emphasize F1 and IoU over OA, which is dominated by the unchanged class under the strong class imbalance of CD masks. On WHU-CD and CDD the strongest baseline (PeftCD) lies within $\le\!0.005$ F1 of BMD-CD, so we regard these results as on par rather than strictly superior; the clearest region-level gains appear on LEVIR-CD (e.g., $+1.4$ F1 over PeftCD) and DSIFN-CD, alongside consistent boundary-F1 improvements in Tab.~\ref{tab:boundary_seed}, suggesting that ordered cross-temporal propagation and logit-space refinement provide complementary benefits. Qualitative results in Fig.~\ref{fig:qualitative} further support these findings: compared to M-CD and DDPM-CD, BMD-CD produces coherent change regions with fewer visible false detections in densely structured scenes.

Table~\ref{tab:zeroshot} reports the challenging in-domain S2Looking evaluation, where BMD-CD obtains 69.23 F1 and 52.94 IoU. It also reports cross-dataset transfer from S2Looking to ValaisCD and B-FLAIR-test without target-domain training, fine-tuning, adaptation, validation, or threshold selection. The threshold is selected on S2Looking
validation and then fixed, as detailed in the supplementary material.

\begin{table}[t]
\caption{Component-wise ablation on LEVIR-CD and WHU-CD.}
\label{tab:ablation_combined}
\centering
\scriptsize
\setlength{\tabcolsep}{4pt}
\renewcommand{\arraystretch}{1.05}

\resizebox{\linewidth}{!}{%
\begin{tabular}{cccc | cccc | cccc}
\toprule
\multirow{2}{*}{Mamba} & \multirow{2}{*}{BOMO} & \multirow{2}{*}{Diff} & \multirow{2}{*}{LTF}
& \multicolumn{4}{c}{LEVIR-CD} 
& \multicolumn{4}{c}{WHU-CD} \\

\cmidrule(lr){5-8} \cmidrule(lr){9-12}
& & & & Prec. & Rec. & F1 & B-F1 
& Prec. & Rec. & F1 & B-F1 \\

\midrule
\checkmark & \xmark  & \xmark & \xmark
& 88.50 & 88.10 & 88.30 & 66.50 
& 94.20 & 93.50 & 93.85 & 80.10 \\

\checkmark & \checkmark & \xmark & \xmark
& 89.15 & 90.95 & 90.04 & 69.10 
& 95.10 & 95.80 & 95.45 & 82.30 \\

\checkmark & \xmark & \xmark & \checkmark
& 89.10 & 89.40 & 89.25 & 72.80
& 94.60 & 94.20 & 94.40 & 84.00 \\

\checkmark & \xmark  & \checkmark & \xmark
& 90.50 & 90.10 & 90.30 & 84.10 
& 95.80 & 94.70 & 95.25 & 88.60 \\

\checkmark & \checkmark & \xmark & \checkmark
& 91.20 & 91.80 & 91.50 & 74.50
& 95.55 & 95.85 & 95.70 & 85.60 \\

\checkmark & \xmark & \checkmark & \checkmark
& 91.45 & 91.10 & 91.27 & 86.20
& 95.75 & 95.15 & 95.45 & 90.10 \\

\midrule
\checkmark & \checkmark & \checkmark & \xmark
& 90.98 & 93.75 & 92.34 & 85.90 
& 94.10 & 94.70 & 94.40 & 89.40 \\

\textbf{\checkmark} & \textbf{\checkmark} & \textbf{\checkmark} & \textbf{\checkmark}
& \textbf{93.41} & \textbf{94.04} & \textbf{93.72} & \textbf{87.70} 
& \textbf{95.80} & \textbf{96.17} & \textbf{95.98} & \textbf{91.40} \\

\bottomrule
\end{tabular}%
}
\end{table}

% --------------------------------------------------------------------------
% 4.7 Ablation Studies
% --------------------------------------------------------------------------
\subsection{Ablation and Analysis}
Table~\ref{tab:ablation_combined} reports a component-wise analysis on LEVIR-CD and WHU-CD. The Mamba-only baseline replaces BOMO with a single bidirectional scan over the flattened, concatenated bi-temporal features (no ordering, LTF, or diffusion); it captures global context but has weak boundaries. Adding BOMO improves recall and region-level F1, indicating that preserving temporal-partition identity in the ordering improves coverage of spatially distributed changes. LTF improves F1 and B-F1 on both datasets with and without diffusion, while diffusion provides the largest single-component boundary gain (e.g., $+8.5$ B-F1 on WHU-CD), confirming that boundary quality is primarily attributable to the logit-space decoder rather than to region-token propagation. The full model achieves the best overall result among the evaluated variants, indicating that BOMO, LTF, and logit-space refinement are most effective when combined. The interaction between ordering and diffusion is not strictly monotonic: on WHU-CD, adding diffusion to the ordered Mamba variant improves B-F1 but reduces region-level F1 when LTF is absent. We attribute this to the logit-space decoder amplifying high-frequency structure: without LTF, $S_{\mathrm{cond}}$ is constructed from less selectively organized deep features, so boundary sharpening can also perturb region interiors and reduce region-level F1.

% --------------------------------------------------------------------------
% 4.8 Graph Construction and Sequence Sensitivity
% --------------------------------------------------------------------------

\subsection{Ordering \& Diffusion Analysis}

Table~\ref{tab:BOMO_sensitivity} isolates token count, ordering, scan direction, and DDIM depth. The default $K=64$, grouped $[a;b]$ order, and bidirectional scan perform best, and five-index DDIM steps saturate the refinement gain. Since random ordering perturbs both spatial and temporal structure, the cleaner comparison is grouped $[a;b]$ vs.\ interleaving, which preserves within-view content and token count yet still loses $1.27$ F1, so the interleaved and random rows act as unstructured-sequence controls that isolate ordering from token count. We attribute the grouped advantage to the sequential scan: processing all $t_1$ tokens before all $t_2$ lets the forward pass consolidate a pre-event context against which post-event tokens are compared, whereas interleaving makes the latent state alternate temporal context at every step. These results support the ordering effect and the five-index accuracy--cost choice.
 
\begin{table}[t]
\caption{BOMO sensitivity analysis on LEVIR-CD. Default: $K\!=\!64$, average pooling, $[a;b]$ ordering, bidirectional scan, and five-index DDIM. The 10-index schedule gives no measurable improvement after rounding.}
\label{tab:BOMO_sensitivity}
\centering
\scriptsize
\setlength{\tabcolsep}{4pt}
\renewcommand{\arraystretch}{1.08}
\begin{tabular}{l|c|c}
\toprule
Configuration & F1 & $\Delta$F1 \\
\midrule
\multicolumn{3}{l}{\textit{Number of bitemporal region nodes $K$}} \\
$K=16$ & 92.35 & $-$1.37 \\
$K=64$ [default] & \textbf{93.72} & --- \\
$K=256$ & 92.88 & $-$0.84 \\
\midrule
\multicolumn{3}{l}{\textit{Sequence ordering}} \\
Interleaved & 92.45 & $-$1.27 \\
$[a;b]$ [default] & \textbf{93.72} & --- \\
Random & 91.80 & $-$1.92 \\
\midrule
\multicolumn{3}{l}{\textit{Scan direction}} \\
Forward & 92.30 & $-$1.42 \\
Bidirectional [default] & \textbf{93.72} & --- \\
\midrule
\multicolumn{3}{l}{\textit{DDIM schedule / denoiser evaluations}} \\
1 denoiser evaluation & 92.62 & $-$1.10 \\
3-index schedule & 92.95 & $-$0.77 \\
5-index schedule [default] & \textbf{93.72} & --- \\
10-index schedule & 93.72 & $<+0.01$ \\
\bottomrule
\end{tabular}
\end{table}

% --------------------------------------------------------------------------
% 4.9 Diffusion vs. Deterministic Refinement
% --------------------------------------------------------------------------
\subsection{Efficiency Analysis}

Table~\ref{tab:efficiency} reports parameters, GFLOPs, and inference latency. For methods with executable implementations, latency is remeasured under a unified NVIDIA RTX 4090 protocol with batch size 1, FP16 inference, synchronized CUDA timing, and $256\times256$ image pairs; methods without executable implementations are reported without latency. The no-diffusion BMD-CD variant requires 13.20 GFLOPs and 22 ms, while the full model requires 32.09 GFLOPs and 47 ms due to the five-index DDIM refinement. This additional cost raises LEVIR-CD F1 from 0.915 to 0.937 and WHU-CD F1 from 0.957 to 0.960. Its main benefit is boundary refinement, as reflected by the diffusion ablations in Table~\ref{tab:ablation_combined} and the B-F1 comparison in Table~\ref{tab:boundary_seed}. Compared with transformer and diffusion-feature baselines, BMD-CD provides competitive accuracy at moderate cost, but PeftCD attains nearly identical WHU-CD F1 with fewer parameters, so the trade-off is dataset-dependent rather than universally Pareto-dominant. Once diffusion is enabled, BMD-CD is also not lower-cost than the lightest Mamba baselines, such as CDMamba at 10.40 GFLOPs; its value is improved accuracy and boundary quality at substantially lower GFLOPs than multi-timestep diffusion-feature extraction, e.g., DDPM-CD at 2175.46 GFLOPs.

\begin{table}[t]
\caption{Computational complexity comparison on LEVIR-CD and WHU-CD. Accuracy values for external methods follow their published reports. Methods without executable implementations are reported without latency.}
\label{tab:efficiency}
\centering
\scriptsize
\setlength{\tabcolsep}{3.2pt}
\renewcommand{\arraystretch}{1.08}
\begin{tabular}{l|c|ccc|cc}
\toprule
Method & Input & Params(M) & GFLOPs & Inf. Time & F1$_{\text{L}}$ & F1$_{\text{W}}$ \\
\midrule

\textbf{Ours (w/o Diff)} 
& 256 & 30.70 & 13.20 & $22^{\dagger}$ ms & 0.915 & 0.957 \\

\textbf{Ours (full)} 
& 256 & 33.55 & 32.09 & $47^{\dagger}$ ms & \textbf{0.937} & \textbf{0.960} \\

\midrule

CDMamba \cite{zhang2024cdmamba} 
& 256 & 11.90 & 10.40 & $15^{\dagger}$ ms & 0.907 & 0.937 \\

PeftCD \cite{dong2026peftcd} 
& 256 & 10.15 & 22.00 & $25^{\dagger}$ ms & 0.923 & 0.959 \\

M-CD \cite{mcd2026} 
& 256 & 69.80 & 29.58 & $160^{\dagger}$ ms & 0.921 & 0.953 \\

LRNet \cite{lrnet2023} 
& 256 & 48.71 & 92.23 & $60^{\dagger}$ ms & 0.911 & 0.925 \\

ChangeFormer \cite{changeformer2022} 
& 256 & 41.02 & 254.80 & $85^{\dagger}$ ms & 0.904 & 0.886 \\

DDPM-CD \cite{bandara2022ddpm} 
& 256 & 46.41 & 2175.46 & -- & 0.909 & 0.927 \\

\midrule

SAM2-CD \cite{qin2025sam2cd} 
& 1024 & 2.59 & 1193.83 & -- & 0.919 & 0.926 \\

\bottomrule
\end{tabular}

\vspace{1mm}
\raggedright
\scriptsize
$^{\dagger}$ Latency is measured from executable implementations under a unified NVIDIA RTX 4090 protocol with batch size 1, FP16 inference, 256×256 image pairs, identical warmup, synchronized CUDA timing, and the same measurement procedure. Accuracy values for external methods are retained from their published papers. Methods without an executable implementation are reported without latency.
\end{table}

\section{Conclusion}

We presented BMD-CD, which combines temporally ordered region-token state-space modeling with logit-space diffusion refinement for remote sensing change detection. BOMO enables efficient long-range cross-temporal interaction, LTF improves change-oriented feature representation, and the diffusion decoder refines coarse predictions for better boundary localization. Experiments on standard benchmarks, S2Looking, and cross-dataset transfer show competitive accuracy with a favorable accuracy--efficiency trade-off, and ablations confirm that temporal ordering, LTF, and logit-space refinement are complementary.

%---------------------------------------------------------------------------------------

{
    \small
    \bibliographystyle{ieeenat_fullname}
    \bibliography{main}
}


\end{document}